\documentclass[runningheads]{llncs}

\usepackage{accv}

\usepackage{accvabbrv}

\usepackage{amsmath}
\usepackage{amssymb}
\usepackage{graphicx}
\usepackage{algorithm}
\usepackage{algpseudocode}
\usepackage{booktabs}
\usepackage{multirow}
\usepackage{makecell}
\usepackage{xcolor}
\usepackage{color, colortbl}
\definecolor{Gray}{gray}{0.9}

\usepackage[accsupp]{axessibility}

\newcommand{\method}{SSCD}
\newcommand{\kl}{\mathrm{KL}}
\newcommand{\simfunc}{\mathrm{sim}}

\newcommand{\stdv}[1]{$\,\scriptstyle\pm #1$}
\usepackage{subcaption}

\usepackage[breaklinks,colorlinks,citecolor=accvblue]{hyperref}

\usepackage{orcidlink}

\usepackage{tikz}
\usetikzlibrary{arrows.meta,positioning,fit}

\begin{document}

\title{Shared Semantic Codebook Distillation for Unpaired Cross-Modal Medical Classification}

\author{
Dillan Imans\inst{1} \and
Phuoc-Nguyen Bui\inst{2} \and
Duc-Tai Le\inst{3} \and
Hyunseung Choo\inst{4}\thanks{Corresponding author}}

\authorrunning{D. Imans et al.}

\institute{
Department of Computer Science and Engineering, \\ Sungkyunkwan University, Suwon, South Korea \and
Convergence Research Institute, \\ Sungkyunkwan University, Suwon, South Korea \and
Department of AI Systems Engineering, \\ Sungkyunkwan University, Suwon, South Korea \and
Department of Electrical and Computer Engineering, \\ Sungkyunkwan University, Suwon, South Korea \\
\email{\{dillanimans,phuocnguyen,ldtai,choo\}@skku.edu}
}

\maketitle

\begin{abstract}
Cross-modal knowledge distillation can transfer diagnostic knowledge from a
strong but costly teacher modality to a cheaper and more deployable student
modality. In medical image analysis, however, the two modalities are often
unpaired: they are collected from different patient cohorts and occupy
geometrically incompatible feature spaces. This makes instance-level
distillation invalid and direct feature matching unreliable. To address these
challenges, we propose Shared Semantic Codebook Distillation (\method{}), which
compares teacher and student representations through a shared discrete codebook.
Each image is represented as a distribution over a common, modality-agnostic
vocabulary, and knowledge is transferred by aligning these distributions across
modalities, both globally and class-conditionally, without requiring paired
samples or directly comparable raw features. The codebook is evolved online by
exponential moving average and kept diverse through entropy regularization and
dead-code restart. At inference, all teacher-side and codebook modules are
discarded, leaving only the student encoder and classifier. On two heterogeneous
unpaired settings, OCT$\rightarrow$fundus retinal disease classification and
CT$\rightarrow$chest-X-ray pneumonia classification, \method{} improves the
student from $64.5$ to $70.2$ macro-F1 and from $73.8$ to $76.3$ macro-F1,
respectively, outperforming all evaluated distillation baselines on both
settings. Code and pretrained models are available at
\url{https://github.com/DillanImans/SSCD-unpaired-distillation}.

\keywords{Cross-modal learning \and Knowledge distillation \and
Discrete codebook learning \and Medical image analysis}

\end{abstract}

\section{Introduction}

Many visual recognition systems are deployed under a modality asymmetry: one
modality carries rich, discriminative information but is costly, invasive, or
unavailable at deployment, while a second modality is cheaper and more widely
accessible but comparatively weak. A natural goal is to let a model that runs
on the accessible modality at test time benefit from the stronger modality
during training. In autonomous driving, LiDAR provides accurate 3D geometry but is an order of magnitude more expensive than cameras, motivating distillation from a LiDAR teacher into a camera-only student that incurs no additional cost
at inference~\cite{wang2023distillbev}. In audio-visual
perception, a visual teacher can supervise a model that must operate from sound
alone when the visual stream is unavailable at test
time~\cite{aytar2016soundnet}. The same asymmetry is pervasive in medical
imaging: computed tomography (CT) offers detailed volumetric structure but is
expensive and exposes patients to higher radiation~\cite{brenner2007computed}, whereas chest X-ray (CXR)
is cheap and ubiquitous in screening; likewise, optical coherence tomography
(OCT) resolves retinal layers that fundus photography cannot~\cite{huang1991optical}, yet fundus
cameras are far more widely deployed. In each case, the goal is the same: transfer knowledge from the strong teacher modality to the weaker student modality that alone is used at deployment. We treat this as a general cross-modal learning problem and study it in the medical setting.

A central obstacle is that, in realistic medical deployments, teacher and
student data are often unpaired: the same patient is not scanned in both
modalities. Paired cross-modal datasets, in which each patient or scene is
captured across both modalities, are scarce, expensive to assemble, and rarely
collected at scale in routine clinical practice. More commonly, the available
data consist of two independently collected cohorts imaged with different
modalities but annotated under a shared label space. This setting breaks
instance-level distillation in two ways. First, there is no valid correspondence
to supervise: a teacher image and a student image sampled in the same mini-batch
usually come from different patients and must not be treated as a matched pair.
Second, even if such a correspondence were artificially imposed, the two
modalities often occupy geometrically incompatible feature spaces because they
differ in appearance, dimensionality, and acquisition geometry. Directly matching
raw features across them is therefore unreliable~\cite{huo2024c2kd}.

Existing approaches do not resolve both difficulties at once. Standard
knowledge distillation assumes that teacher and student share the same input or
operate in directly comparable representation spaces~\cite{hinton2015distilling,romero2015fitnetshintsdeepnets,park2019relational,tian2019contrastive, zagoruyko2016paying},
which fails once the modalities differ substantially. Paired cross-modal
distillation relaxes the shared-input assumption by transferring supervision
across modalities, but still requires paired, registered, or synchronized
samples to define the correspondence~\cite{gupta2016cross,aytar2016soundnet,wang2023distillbev}. These methods
therefore become inapplicable when teacher and student datasets are collected
independently. Recent unpaired medical distillation methods address the absence
of paired samples~\cite{wang2023fundus,wang2024multieye}, but they either align
teacher and student through continuous feature or prediction statistics, which
implicitly assumes that the two embedding spaces can be made comparable, or
route distillation through an externally defined semantic medium, which
introduces additional modality- or domain-specific dependencies. What remains
open is a generic framework that can handle both unpaired data and incompatible
feature spaces without relying on paired correspondence, direct feature-space
comparability, or external modality-specific priors.

To address this, we propose \textbf{Shared Semantic Codebook Distillation} (\method{}). The core idea is to avoid comparing teacher and student features directly and instead route both modalities through a shared discrete semantic codebook that acts as a neutral intermediate space. Teacher and student embeddings are softly assigned to the same set of learned codes, so each image is represented not by its raw features but by a distribution over a common semantic vocabulary. Because both modalities are expressed in this shared
codebook space, \method{} sidesteps the geometric incompatibility that makes
direct cross-modal feature matching unreliable. To respect the unpaired setting, \method{} matches teacher and student code-assignment distributions rather than individual samples: it aligns their overall code usage and crucially their class-conditional code distributions, so the average code signature of each disease agrees across modalities without ever assuming an instance-level correspondence. The distillation mechanism is modality-agnostic and can be applied unchanged across different modality pairs. At inference, all teacher-side
and codebook modules are discarded, leaving only the student encoder and classifier, so \method{} adds no deployment cost over the student-only model.

Our contributions are summarized as follows:
\begin{itemize}
    \item We study unpaired cross-modal medical image classification under a
    practical training-time teacher and test-time student setting, where the two
    modalities share labels but not patient-level correspondence.

    \item We propose \method{}, which uses a shared semantic codebook to bridge
    heterogeneous teacher and student representations. This converts
    modality-specific features into a common code vocabulary before distillation.

    \item We design a code-distribution distillation objective that transfers
    both global modality-level knowledge and class-level disease structure,
    enabling each class to be described by multiple semantic codes rather than a
    single prototype.

    \item We validate \method{} on OCT$\rightarrow$fundus and CT$\rightarrow$CXR
    classification, showing consistent improvements over the student-only model
    and all evaluated distillation baselines with no additional inference cost.
\end{itemize}

\section{Related Work}

\subsection{Distillation}
\label{sec:distillation}

Knowledge distillation transfers knowledge from a high-capacity teacher to a
student model. Early work used softened output logits~\cite{hinton2015distilling},
while later methods distilled intermediate feature hints~\cite{romero2015fitnetshintsdeepnets}, relational structure~\cite{park2019relational}, and contrastive representations~\cite{tian2019contrastive}. These methods are effective when teacher and
student operate in the same modality or have directly comparable
representations, but this assumption becomes fragile when modalities differ in
appearance, dimensionality, and feature geometry~\cite{huo2024c2kd}.

Cross-modal distillation relaxes the shared-input assumption by transferring
knowledge from a richer training-time modality to another modality used at
deployment. Prior work has transferred supervision from RGB to depth using
registered image pairs~\cite{gupta2016cross}, from vision to sound using
synchronized audio-visual data~\cite{aytar2016soundnet}, and from LiDAR to
camera using shared driving scenes~\cite{wang2023distillbev}. However, these
methods still rely on paired, registered, or synchronized samples, which are
often unavailable in medical imaging.

Recent unpaired medical distillation methods address this limitation.
FDDM~\cite{wang2023fundus} aligns OCT and fundus images through class-prototype
matching and class-similarity alignment in a continuous feature space, while
OCT-CoDA~\cite{wang2024multieye} uses language-model-generated disease concepts
and vision-language image--concept similarities as an external semantic bridge.
In contrast, \method{} learns a shared discrete codebook directly from the
teacher-student training process. Each image is represented as a distribution
over reusable semantic codes, and distillation matches global and
class-conditional code distributions rather than continuous prototypes or
externally defined concepts. This provides a modality-agnostic bridge for
unpaired cross-modal distillation without paired samples or direct feature-space
matching.

\subsection{Codebooks and Assignment-Based Representation Learning}

Codebook-based representation learning maps continuous features to discrete or
prototype-like entries in a learned vocabulary. VQ-VAE introduced discrete
latent codes updated from encoded features~\cite{van2017neural,esser2021taming}, while
BEiT~v2 used vector-quantized distillation to produce semantic visual tokens
~\cite{peng2022beit}. Assignment-based self-supervised methods such as DeepCluster~\cite{caron2018deep}, SwAV~\cite{caron2020unsupervised}, and DINO~\cite{caron2021emerging} further
show that matching cluster or prototype assignments can learn strong semantic
representations. More recently, VQGraph used soft code assignments for
distillation between different model families~\cite{yang2024vqgraph}. These works motivate our use of assignment distributions, but they mainly use
codebooks within a single modality or for same-domain transfer. In contrast,
\method{} uses a shared codebook as a cross-modal bridge for unpaired datasets:
teacher and student modalities are expressed as distributions over one common
vocabulary, and knowledge is transferred by aligning their global and
class-conditional code distributions.


\section{Proposed Method}

\subsection{Problem Formulation}

We consider unpaired cross-modal knowledge distillation between a teacher
modality and a student modality. Let
$D_T=\{(x_T^i,y_T^i)\}_{i=1}^{N_T}$ denote a labeled teacher-modality dataset and
$D_S=\{(x_S^j,y_S^j)\}_{j=1}^{N_S}$ denote a labeled student-modality dataset. The
two datasets share the same label space $\mathcal{Y}$, but they are not paired: a
teacher image $x_T^i$ and a student image $x_S^j$ do not correspond to the same
patient or clinical acquisition. This setting is common in medical imaging, where
datasets from different modalities are often collected independently.

A teacher encoder $f_T$ is trained on the teacher modality and frozen during
distillation. The goal is to train a student model, composed of an encoder
$f_S$ and classifier $\phi_S$, that is deployed using only the student modality. Since teacher
and student images are unpaired and visually heterogeneous, direct
instance-level feature matching is ill-defined. We therefore formulate
cross-modal distillation as semantic distribution matching over a shared
codebook space, both overall and class-conditionally.

\subsection{Overview}

\begin{figure}[t]
\centering
\includegraphics[width=\linewidth]{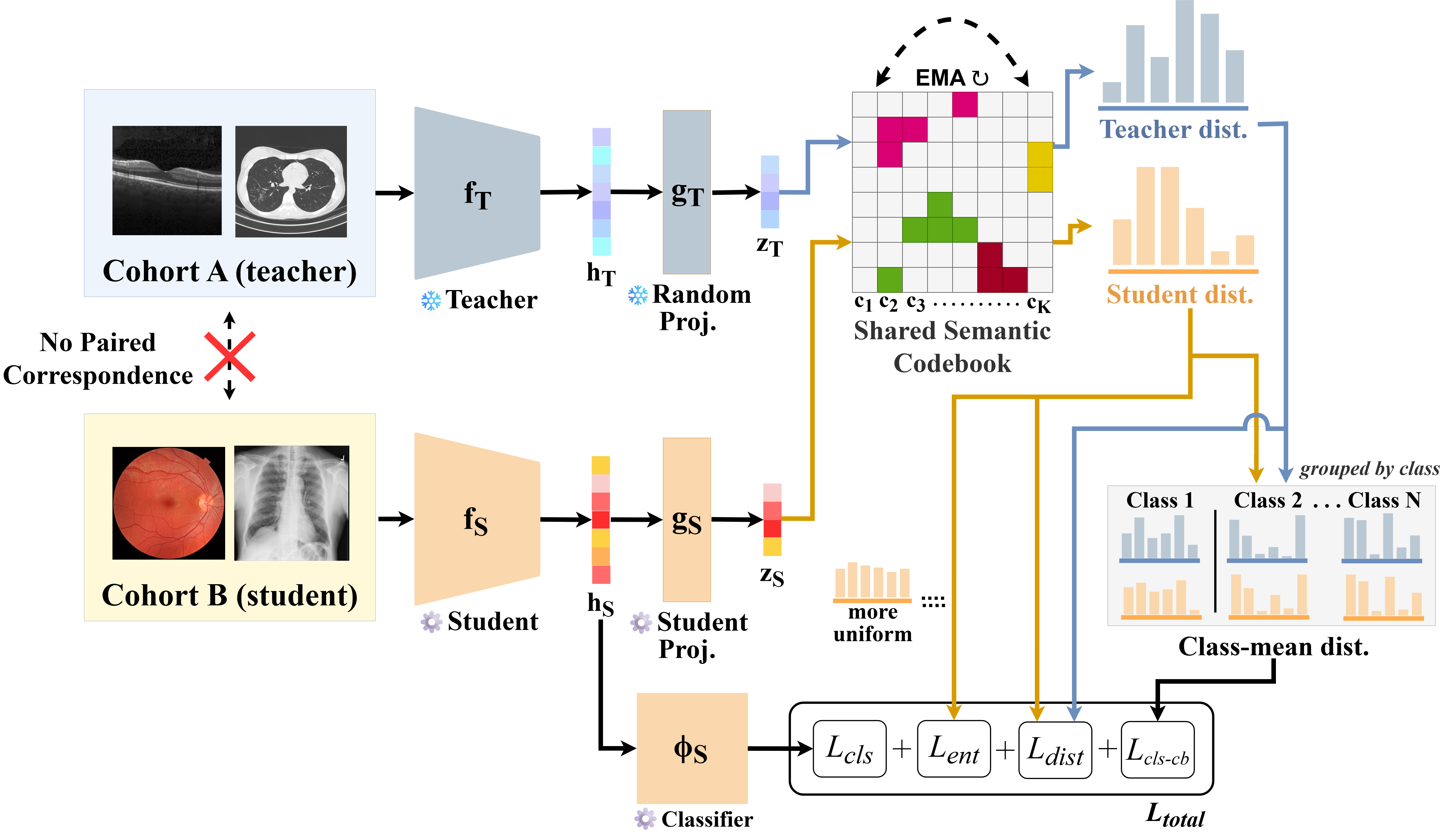}
\caption{Overview of \method{}. The two images shown per cohort illustrate
the two settings evaluated (OCT$\rightarrow$fundus and CT$\rightarrow$CXR),
each trained independently with a single teacher and single student
modality. Teacher and student features are projected into a shared space
and softly assigned to a common codebook, where their code distributions
are aligned overall and per class.}
\label{fig:method}
\end{figure}

Figure~\ref{fig:method} gives an overview of \method{}. It routes both modalities through a shared discrete codebook that acts as a neutral intermediate space, so that the teacher and student are never compared in their raw feature geometries. Each image is represented not by its features
but by a soft distribution over a common set of learned codes, and knowledge is
transferred by aligning these distributions across modalities. The framework has
four parts, described in turn below: (i) a projection of each backbone's features
into a shared space (Sec.~\ref{sec:proj}); (ii) a shared semantic codebook with a
soft code-assignment rule (Sec.~\ref{sec:codebook}); (iii) a cross-modal
distillation objective that matches code distributions both globally and
class-conditionally (Sec.~\ref{sec:distill}); and (iv) an online EMA evolution of
the codebook with diversity regularization (Sec.~\ref{sec:evolve}). The codebook,
the projectors, and the teacher are all training-time scaffolding: at inference,
they are discarded and only the student encoder and classifier run
(Sec.~\ref{sec:train_infer}), so \method{} adds no cost over a plain student model.

\subsection{Projecting Modalities into the Shared Space}
\label{sec:proj}

Given a teacher image $x_T$ and a student image $x_S$, the corresponding backbones produce a
pooled feature
\begin{equation}
  h_T = f_T(x_T), \qquad h_S = f_S(x_S), \qquad h_T, h_S \in \mathbb{R}^{D}.
\end{equation}
Because $h_T$ and $h_S$ originate from different modalities, they are not
directly comparable. We therefore map each into a shared, lower-dimensional
space with modality-specific projection heads,
\begin{equation}
  z_T = g_T(h_T), \qquad z_S = g_S(h_S), \qquad z_T, z_S \in \mathbb{R}^{d},
\end{equation}
each a small two-layer MLP (architecture details in Sec.~\ref{sec:setup}). The
projected embeddings $z_T,z_S$ are used only for codebook assignment; the
student classifier operates separately on the raw backbone feature,
$\phi_S(h_S)$. The classification head and the codebook head are thus two
distinct branches off the same student backbone, reading the $D$- and
$d$-dimensional representations respectively.

\paragraph{Frozen teacher projection.}
The student projector $g_S$ is trained, but the teacher projector $g_T$ stays
fixed at its random initialization for the entire run, for two reasons. First,
the codebook evolves by a slow EMA (Sec.~\ref{sec:evolve}) and can only form
clean clusters if the distribution it tracks is stationary; a trained $g_T$
would make the teacher side a moving target, whereas a frozen $g_T$ gives a
fixed teacher manifold. Second, we expect the class structure of the
high-accuracy frozen teacher backbone to largely survive an untrained
projection: although $g_T$ is a shallow nonlinear MLP rather than a strictly
linear map, random projections are known to approximately preserve the
geometry of well-organized representations in the spirit of the
Johnson--Lindenstrauss phenomenon~\cite{johnson1984extensions}, and we find
this holds empirically, as the teacher's class separability in
Fig.~\ref{fig:both} and the cross-modal alignment in Table~\ref{tab:codebook}
both indicate. The
transferable knowledge already lives in the trained teacher backbone, and the
projector's only job is dimensionality reduction, not representation learning.

\subsection{Shared Semantic Codebook and Soft Assignment}
\label{sec:codebook}

The shared codebook is a set of $K$ code vectors,
\begin{equation}
  C=\{c_k\}_{k=1}^{K}, \qquad c_k \in \mathbb{R}^{d}.
\end{equation}
Each code acts as a semantic anchor in the shared space. The same codebook is
used for both modalities, so teacher and student images are expressed using one
common vocabulary. The codebook is maintained as a buffer and is updated by the
EMA rule described in Sec.~\ref{sec:evolve}, rather than by gradient descent.

For a projected embedding $z$, we compute a soft assignment over the codes using
temperature-scaled cosine similarity,
\begin{equation}
  p(k \mid z)=
  \frac{\exp\!\big(\simfunc(z,c_k)/\tau\big)}
       {\sum_{l=1}^{K}\exp\!\big(\simfunc(z,c_l)/\tau\big)},
  \label{eq:softassign}
\end{equation}
where $\simfunc(\cdot,\cdot)$ is cosine similarity and $\tau$ is a temperature
that sharpens the assignment as it decreases. The resulting vector
$p(\cdot\mid z)\in\mathbb{R}^{K}$ is a probability distribution over codes. We
write the teacher and student assignments as
\begin{equation}
  p_T = p(\cdot \mid z_T), \qquad p_S = p(\cdot \mid z_S).
\end{equation}
These distributions are a modality-agnostic representation of each image: both
modalities are described in the same vocabulary, so they can be compared even
though their raw features cannot. Gradients from the distillation losses flow
through $z_S$ into $g_S$ and the student backbone via Eq.~\eqref{eq:softassign}; the
codes $c_k$ themselves receive no gradient.

\subsection{Cross-Modal Distillation Objective}
\label{sec:distill}

We transfer knowledge by aligning the teacher and student code distributions with
two complementary terms. In both, the teacher distribution is treated as a fixed
target and detached from the gradient computation.

\paragraph{Distribution distillation.}
The first term pulls the student's batch-marginal code distribution toward the
teacher's under the forward KL divergence. Let
\begin{equation}
  \bar{p}_T = \frac{1}{B_T}\sum_{i=1}^{B_T} p_T^{(i)},
  \qquad
  \bar{p}_S = \frac{1}{B_S}\sum_{j=1}^{B_S} p_S^{(j)}
\end{equation}
denote the teacher and student marginal code distributions over the current
mini-batches, aligned with
\begin{equation}
  \mathcal{L}_{\mathrm{dist}}
  =
  \kl\!\big(\,\bar{p}_T \,\big\|\, \bar{p}_S\,\big).
  \label{eq:ldist}
\end{equation}
Because the teacher is the reference distribution, the student is penalized for
placing low probability, in aggregate, on codes the teacher activates, so it
learns to use the same regions of the shared vocabulary as the teacher without
any instance-level correspondence.

\paragraph{Class-conditional codebook alignment.}
Since the data are unpaired, the principal transfer signal is label-conditioned
rather than instance-level. For each class $c$ present with at least two
samples in both the teacher and student mini-batches, we form the class-mean
code distributions
\begin{equation}
  \bar{p}_T^{\,c}
  =
  \frac{1}{N_T^c}\!\!\sum_{i:\,y_T^i=c}\!\! p_T^{(i)},
  \qquad
  \bar{p}_S^{\,c}
  =
  \frac{1}{N_S^c}\!\!\sum_{j:\,y_S^j=c}\!\! p_S^{(j)},
  \label{eq:classmean}
\end{equation}
where $N_T^c$ and $N_S^c$ count the class-$c$ samples in each batch, aligned
with a squared-error loss,
\begin{equation}
  \mathcal{L}_{\mathrm{cls\text{-}cb}}
  =
  \frac{1}{|\mathcal{Y}_B|}
  \sum_{c\in\mathcal{Y}_B}
  \big\|\, \bar{p}_S^{\,c} - \bar{p}_T^{\,c} \,\big\|_2^2,
  \label{eq:lclasscb}
\end{equation}
where $\mathcal{Y}_B$ is the set of classes satisfying this condition in the
current step. This term asks the teacher's and student's average
code-signature for each disease to agree; classes absent from either batch are
skipped for that step.

\subsection{Codebook Evolution and Diversity Regularization}
\label{sec:evolve}

\paragraph{EMA update.}
After each optimizer step, the codebook is refreshed using the freshly
computed projections of both modalities,
$\mathcal{B}=\{z_T^i\}_{i=1}^{B_T}\cup\{z_S^j\}_{j=1}^{B_S}$; this is a
separate buffer update, outside the backpropagation graph entirely, so it has
no effect on the trained network's gradients. The losses use the soft
assignment of Eq.~\eqref{eq:softassign}; the EMA instead uses a hard top-1
assignment, $\kappa(z)=\arg\max_k \simfunc(z,c_k)$, meaning each embedding
votes for its single closest code. We keep a running count $n_k$ (how many
embeddings have recently landed on code $k$) and a running sum $m_k$ (their
accumulated direction), updated each step as an exponential moving average
with momentum $\gamma$:
\begin{align}
  n_k &\leftarrow \gamma\, n_k + (1-\gamma)\!\!\sum_{z\in\mathcal{B}}\!\!
        \mathbb{1}\!\left[\kappa(z)=k\right], \\
  m_k &\leftarrow \gamma\, m_k + (1-\gamma)\!\!\sum_{z\in\mathcal{B}}\!\!
        \mathbb{1}\!\left[\kappa(z)=k\right]\,\hat{z},
\end{align}
where $\hat{z}=z/\|z\|$ is the unit-normalized embedding. The code vector is
then the (Laplace-smoothed) running mean of the embeddings assigned to it,
renormalized to unit length:
\begin{equation}
  c_k \leftarrow \mathrm{normalize}\!\left(\frac{m_k}{n_k+\epsilon}\right).
  \label{eq:emacode}
\end{equation}
This is simply an online, EMA-smoothed $k$-means: each code slowly drifts
toward the centroid of the embeddings that land on it. Pooling embeddings
from both modalities keeps codes distributed across the teacher's region and
the shifting student's region, so neither side is left without nearby codes.

\paragraph{Dead-code restart.}
EMA $k$-means can leave codes unused. Periodically, any code whose count
falls below a small threshold is re-seeded to a randomly chosen embedding
from the current batch and its statistics are reset, which keeps the full
vocabulary alive and prevents collapse onto a few dominant codes. This is
standard practice in VQ-based training~\cite{van2017neural} and triggers
rarely under our settings, so we do not ablate it separately from
$\mathcal{L}_{\mathrm{ent}}$, the diversity term we do ablate.

\paragraph{Diversity regularization.}
As a second, gradient-based guard against collapse, we maximize the entropy
of the batch-averaged student assignment. Let
$\bar{p}_S = \frac{1}{B_S}\sum_{j=1}^{B_S} p_S^{(j)}$ be the student's
marginal code usage over the batch; we minimize its negative entropy,
\begin{equation}
  \mathcal{L}_{\mathrm{ent}}
  =
  -H(\bar{p}_S)
  =
  \sum_{k=1}^{K} \bar{p}_S(k)\,\log \bar{p}_S(k).
  \label{eq:lent}
\end{equation}
This pushes marginal code usage toward uniform, spreading aggregate code usage across the vocabulary and preventing collapse onto a few dominant codes. Like $\mathcal{L}_{\mathrm{dist}}$, it acts on the batch marginal rather than on individual assignments, and it is applied to the student only.

\subsection{Training and Inference}
\label{sec:train_infer}

\paragraph{Objective.}
The student is trained with the standard classification loss and the three
codebook terms,
\begin{equation}
  \mathcal{L}
  =
  \mathcal{L}_{\mathrm{cls}}
  + \lambda_{\mathrm{dist}}\,\mathcal{L}_{\mathrm{dist}}
  + \lambda_{\mathrm{cls\text{-}cb}}\,\mathcal{L}_{\mathrm{cls\text{-}cb}}
  + \lambda_{\mathrm{ent}}\,\mathcal{L}_{\mathrm{ent}},
  \label{eq:total}
\end{equation}
where $\mathcal{L}_{\mathrm{cls}}$ is class-weighted cross-entropy on the
student logits $\phi_S(h_S)$, and the $\lambda$ terms balance the codebook
losses against classification (values in Sec.~\ref{sec:setup}). Note that $\mathcal{L}_{\mathrm{ent}}$ pulls the student marginal toward uniform while $\mathcal{L}_{\mathrm{dist}}$ pulls it toward the teacher marginal $\bar{p}_T$; we keep $\lambda_{\mathrm{ent}}$ small so that $\mathcal{L}_{\mathrm{ent}}$ acts only as a guard against marginal collapse rather than competing with $\mathcal{L}_{\mathrm{dist}}$ where the teacher signal is informative. Each training
step draws one teacher and one student mini-batch independently; the teacher
loader is cycled if shorter.

\paragraph{Optimization.}
The student encoder $f_S$, classifier $\phi_S$, and student projector $g_S$ are
optimized end to end. The teacher encoder $f_T$ is frozen throughout
distillation, and the codebook $C$ is updated only through EMA. Training begins
with a short warmup stage in which only $\mathcal{L}_{\mathrm{cls}}$ is active
while the codebook EMA continues to run. This allows the codebook to settle on
the feature distribution before the distillation losses are introduced. The
checkpoint with the best validation macro-F1 is selected for evaluation.

\paragraph{Inference.}
At test time, all distillation modules are removed. \method{} introduces no additional inference cost over the student-only baseline. Given a student-modality image $x_S$, prediction uses only the student encoder and classifier:
\begin{equation}
  \hat{y} = \arg\max_y\ \phi_S(f_S(x_S)).
\end{equation}

\section{Experiments}

\subsection{Experimental Setup}
\label{sec:setup}
\paragraph{Datasets.}

We evaluate \method{} on two unpaired cross-modal medical classification
settings that differ in imaging dimensionality, anatomy, and class count.
\textbf{OCT$\rightarrow$fundus} uses MultiEYE~\cite{wang2024multieye}, where OCT
serves as the teacher modality and fundus photography as the student modality.
Following our unpaired setting, the two modalities share the same disease label
space but contain different patient cohorts. We use six categories with
sufficient samples in both modalities: normal, dry age-related macular
degeneration (dAMD), diabetic retinopathy (DR), macular epiretinal membrane
(MEM), retinal vein occlusion (RVO), and wet age-related macular degeneration
(wAMD). The remaining three MultiEYE categories are excluded because one modality
contains too few samples; the same six-class split is used for all methods.
Fundus images are enhanced with CLAHE, OCT images are median-filtered, and all
images are resized to $224\times224$.

\textbf{CT$\rightarrow$CXR} evaluates a more heterogeneous cross-anatomy setting.
We use COVIDx CT-3A~\cite{gunraj2022covidx} as the CT teacher dataset and NIH
ChestX-ray~\cite{wang2017chestx} as the CXR student dataset. COVIDx CT-3A is
binarized into normal and pneumonia. NIH ChestX-ray
is mapped to the same nominal label space by assigning ``No Finding'' to normal
and ``Pneumonia'' to pneumonia, while excluding other categories; pneumonia
images with co-occurring findings are retained. This setting is fully unpaired:
the teacher and student images come from independently collected cohorts and
share only a normal-versus-pneumonia label space, rather than patient-level
correspondence or clinically identical label definitions.

\paragraph{Baselines.}
We compare with a student-only model, two simple mean-matching baselines, and
FDDM~\cite{wang2023fundus}. DistillGlobMean aligns the student's projected
embedding to a single global teacher mean, while DistillClsMean aligns it to the
corresponding class-wise teacher mean using MSE. FDDM is the closest unpaired
cross-modal medical distillation baseline and is reimplemented under the same
backbone, optimizer, and data protocol for both settings. We exclude
OCT-CoDA~\cite{wang2024multieye} from the main comparison because it requires
LLM-generated disease concepts and a pretrained vision-language backbone, which
breaks the common ResNet-50 protocol and is not directly transferable to the
CT$\rightarrow$CXR setting.

\paragraph{Implementation.}
All methods use an ImageNet-pretrained ResNet-50 backbone with
$224\times224$ inputs and are trained under the same protocol unless otherwise
specified. We use AdamW with learning rate $3\times10^{-5}$, weight decay
$0.01$, and a cosine learning-rate schedule. Models are trained for up to
$50$ epochs with batch size $32$, and early stopping is applied with patience
$15$ based on validation macro-F1. The student classification objective is
inverse-frequency class-weighted cross-entropy. For each run, the checkpoint
with the best validation macro-F1 is evaluated once on the held-out test set.

For \method{}, the ResNet-50 pooled feature dimension is $D=2048$. The teacher
and student projection heads map features to a $d=128$ codebook space using a
two-layer MLP (Linear$\rightarrow$ReLU$\rightarrow$Linear). We use $K=256$
codes, assignment temperature $\tau=0.1$,
$\lambda_{\mathrm{dist}}=\lambda_{\mathrm{cls\text{-}cb}}=0.5$, and
$\lambda_{\mathrm{ent}}=0.05$. The codebook is updated with EMA momentum
$\gamma=0.99$, and inactive codes are re-seeded when their count falls below
$\eta=1$ every $R=100$ steps. During the first $3$ epochs, codebook losses are
disabled while the EMA update remains active, allowing the codebook to stabilize
before distillation begins.

FDDM is implemented with the same backbone, optimizer, and training schedule,
using only its original loss weights. DistillGlobMean and DistillClsMean sweep
$\lambda\in\{0.05,0.1,0.3,0.5,1\}$ and select the value that maximizes average
validation macro-F1 across the two modality pairs; this gives $\lambda=0.5$ for
DistillGlobMean and $\lambda=1.0$ for DistillClsMean. Full sweep results are
provided in the supplementary material.

\paragraph{Evaluation metrics.}
We report precision, recall, F1, AUC, and Cohen's $\kappa$ as mean$\pm$std over
three seeds, with macro-F1 as the primary metric. For each seed, we regenerate
the patient-level train/validation/test split and model initialization, so the
reported variation reflects both data-partition and optimization randomness.

\subsection{Main Results}
\label{sec:main}

\begin{table}[t]
\centering
\setlength{\tabcolsep}{6pt}
\caption{OCT$\rightarrow$fundus multiclass retinal disease classification on
MultiEYE (6 classes). OCT is the teacher modality, fundus the student.
Mean $\pm$ std over three seeds; best deployable result in \textbf{bold}. The OCT teacher is an upper-bound reference.}
\label{tab:multieye}
\begin{tabular}{lccccc}
\toprule
Method & F1 & Prec. & Rec. & AUC & $\kappa$ \\
\midrule
OCT teacher        & 88.0\stdv{2.6} & 86.8\stdv{2.5} & 90.4\stdv{1.9} & 98.7\stdv{0.4} & 88.7\stdv{1.6} \\
\midrule
Fundus student     & 64.5\stdv{3.0} & 74.7\stdv{1.5} & 61.0\stdv{2.8} & 90.2\stdv{0.7} & 46.6\stdv{3.2} \\
DistillGlobMean    & 66.3\stdv{1.9} & 78.2\stdv{3.2} & 62.4\stdv{0.9} & 90.8\stdv{0.3} & 48.0\stdv{2.2} \\
DistillClsMean     & 65.5\stdv{2.1} & 77.6\stdv{3.3} & 61.9\stdv{0.9} & \textbf{91.2}\stdv{0.2} & 47.6\stdv{2.8} \\
FDDM~\cite{wang2023fundus} & 66.2\stdv{2.6} & 78.0\stdv{0.7} & 62.1\stdv{2.5} & 90.0\stdv{0.8} & 46.7\stdv{1.9} \\
\rowcolor{Gray}
\textbf{\method{} (Ours)} & \textbf{70.2}\stdv{1.6} & \textbf{79.9}\stdv{2.3} & \textbf{65.3}\stdv{1.5} & 90.7\stdv{0.3} & \textbf{52.1}\stdv{1.8} \\
\bottomrule
\end{tabular}
\end{table}

\begin{table}[t]
\centering
\setlength{\tabcolsep}{6pt}
\caption{CT$\rightarrow$CXR binary normal-versus-pneumonia classification
(COVIDx CT-3A teacher, NIH ChestX-ray student). Mean $\pm$ std over three
seeds; best deployable result in \textbf{bold}. The CT teacher is an
upper-bound reference.}
\label{tab:ctxr}
\begin{tabular}{lccccc}
\toprule
Method & F1 & Prec. & Rec. & AUC & $\kappa$ \\
\midrule
CT teacher         & 98.9\stdv{0.1} & 98.9\stdv{0.2} & 98.9\stdv{0.1} & 99.9\stdv{0.1} & 97.8\stdv{0.2} \\
\midrule
X-ray student      & 73.8\stdv{0.9} & 73.5\stdv{0.9} & 75.0\stdv{1.0} & 80.8\stdv{0.7} & 47.9\stdv{1.8} \\
DistillGlobMean    & 72.8\stdv{0.7} & 72.5\stdv{0.6} & 74.0\stdv{0.6} & 81.1\stdv{0.5} & 45.9\stdv{1.2} \\
DistillClsMean     & 74.7\stdv{1.5} & 74.5\stdv{1.3} & 76.3\stdv{1.2} & 81.2\stdv{0.4} & 49.9\stdv{2.8} \\
FDDM~\cite{wang2023fundus} & 74.4\stdv{1.9} & 74.2\stdv{1.9} & 76.0\stdv{2.1} & 81.7\stdv{1.2} & 49.3\stdv{3.8} \\
\rowcolor{Gray}
\textbf{\method{} (Ours)} & \textbf{76.3}\stdv{1.3} & \textbf{75.9}\stdv{1.1} & \textbf{77.4}\stdv{0.8} & \textbf{82.9}\stdv{1.5} & \textbf{52.7}\stdv{2.4} \\
\bottomrule
\end{tabular}
\end{table}

Tables~\ref{tab:multieye} and~\ref{tab:ctxr} report both settings.
\method{} achieves the best macro-F1 among all methods on both settings,
improving over the student-only baseline by $+5.7$ on
OCT$\rightarrow$fundus and $+2.5$ on CT$\rightarrow$CXR, and outperforming
FDDM and the other distillation baselines. Precision, recall, and $\kappa$
move together with F1 in both tables. AUC is largely saturated across
methods on the six-class pair, where all methods including the student-only
baseline fall within a narrow band ($90.0$--$91.2$); \method{} is
competitive but not the highest on this metric there, while remaining best
on AUC on the binary pair.
The naive mean-transfer baselines are inconsistent across settings: modest
gains comparable to FDDM on OCT$\rightarrow$fundus, but on
CT$\rightarrow$CXR DistillGlobMean falls below the student while
DistillClsMean edges past FDDM, showing that matching averaged teacher
statistics alone is not a reliable strategy across modality pairs.
\method{} is the only method that improves the student consistently on both
settings and across nearly all metrics.

\subsection{Ablation Study}
\label{sec:ablation}

\begin{table}[t]
\centering
\setlength{\tabcolsep}{6pt}
\caption{Ablation of the three codebook losses, removing one term at a time from
the full \method{} objective. Mean $\pm$ std over three seeds.}
\label{tab:ablation}
\begin{tabular}{lccccc}
\toprule
Variant & F1 & Prec. & Rec. & AUC & $\kappa$ \\
\midrule
\multicolumn{6}{c}{\emph{OCT$\rightarrow$fundus (6 classes)}}\\
\midrule
\rowcolor{Gray}
\method{} (full) & \textbf{70.2}\stdv{1.6} & \textbf{79.9}\stdv{2.3} & \textbf{65.3}\stdv{1.5} & 90.7\stdv{0.3} & \textbf{52.1}\stdv{1.8} \\
\quad $-\,\mathcal{L}_{\mathrm{dist}}$       & 68.6\stdv{2.1} & 79.2\stdv{1.9} & 63.6\stdv{1.8} & 90.2\stdv{0.4} & 48.9\stdv{0.8} \\
\quad $-\,\mathcal{L}_{\mathrm{cls\text{-}cb}}$ & 69.3\stdv{1.2} & 79.0\stdv{1.4} & 64.6\stdv{1.5} & 90.7\stdv{0.3} & 50.6\stdv{0.8} \\
\quad $-\,\mathcal{L}_{\mathrm{ent}}$        & 68.2\stdv{1.8} & 78.0\stdv{0.7} & 63.8\stdv{1.3} & 90.4\stdv{0.3} & 48.5\stdv{1.6} \\
\midrule
\multicolumn{6}{c}{\emph{CT$\rightarrow$CXR (2 classes)}}\\
\midrule
\rowcolor{Gray}
\method{} (full) & 76.3\stdv{1.3} & 75.9\stdv{1.1} & 77.4\stdv{0.8} & 82.9\stdv{1.5} & 52.7\stdv{2.4} \\
\quad $-\,\mathcal{L}_{\mathrm{dist}}$       & 74.4\stdv{3.5} & 74.2\stdv{3.5} & 75.4\stdv{3.5} & 80.9\stdv{3.0} & 49.1\stdv{6.9} \\
\quad $-\,\mathcal{L}_{\mathrm{cls\text{-}cb}}$ & 75.8\stdv{1.4} & 75.5\stdv{1.3} & 77.0\stdv{0.9} & 83.9\stdv{0.9} & 51.9\stdv{2.7} \\
\quad $-\,\mathcal{L}_{\mathrm{ent}}$        & \textbf{76.6}\stdv{1.0} & \textbf{76.2}\stdv{1.0} & 77.4\stdv{0.6} & \textbf{84.4}\stdv{0.5} & \textbf{53.4}\stdv{1.9} \\
\bottomrule
\end{tabular}
\end{table}

\begin{table}[!h]
\centering
\setlength{\tabcolsep}{6pt}
\caption{Cross-modal alignment in codebook space: per-class cosine similarity
between teacher and student class-mean code distributions, with per-class
teacher ($n_T$) and student ($n_S$) sample counts. Mean $\pm$ std over three
seeds.}
\label{tab:codebook}
\setlength{\tabcolsep}{9pt}
\begin{tabular}{lccc}
\toprule
Class & $n_T$ & $n_S$ & Cosine sim. \\
\midrule
\multicolumn{4}{c}{OCT$\rightarrow$fundus (6 classes), mean $=0.706$}\\
\midrule
Normal & 432.3\stdv{53.6}  & 450.0\stdv{17.7} & 0.759\stdv{0.042} \\
DR     & 450.7\stdv{122.4} & 445.0\stdv{6.2}  & 0.765\stdv{0.076} \\
dAMD   & 28.0\stdv{2.8}    & 204.3\stdv{5.6}  & 0.799\stdv{0.187} \\
MEM    & 149.3\stdv{30.9}  & 29.0\stdv{1.4}   & 0.662\stdv{0.081} \\
RVO    & 25.3\stdv{2.5}    & 66.7\stdv{1.2}   & 0.652\stdv{0.293} \\
wAMD   & 150.3\stdv{35.9}  & 62.7\stdv{0.9}   & 0.600\stdv{0.114} \\
\midrule
\multicolumn{4}{c}{CT$\rightarrow$CXR (2 classes), mean $=0.519$}\\
\midrule
Normal    & 744.7\stdv{20.5} & 200.0\stdv{0.0} & 0.490\stdv{0.142} \\
Pneumonia & 714.7\stdv{38.0} & 109.7\stdv{2.6} & 0.547\stdv{0.095} \\
\bottomrule
\end{tabular}
\end{table}

\begin{figure}[h]
    \centering
    \begin{subfigure}{0.33\linewidth}
        \centering
        \includegraphics[width=\linewidth]{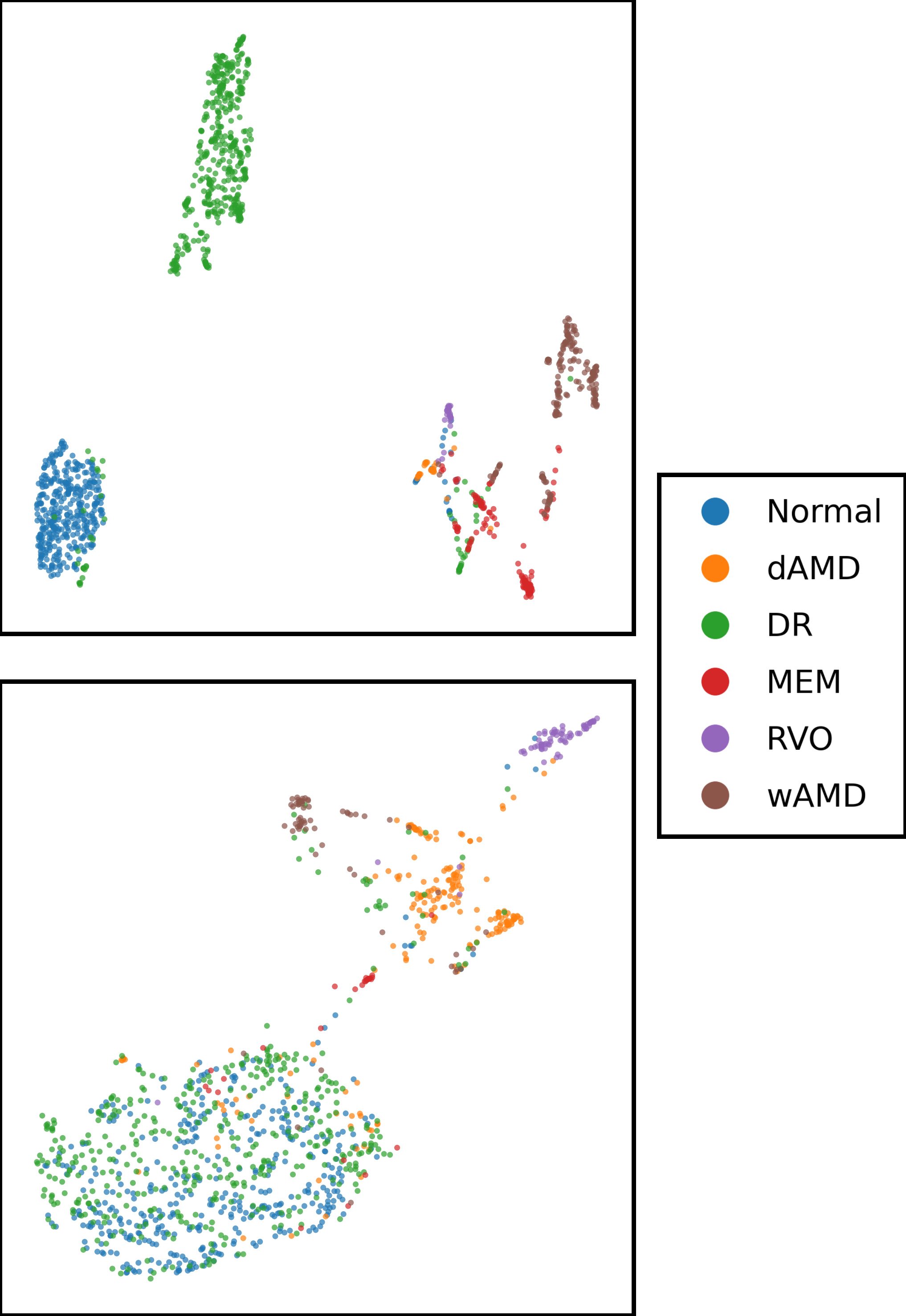}
        \caption{OCT$\rightarrow$Fundus}
        \label{fig:multieye}
    \end{subfigure}
    \hspace{0.02\linewidth}
    \begin{subfigure}{0.36\linewidth}
        \centering
        \includegraphics[width=\linewidth]{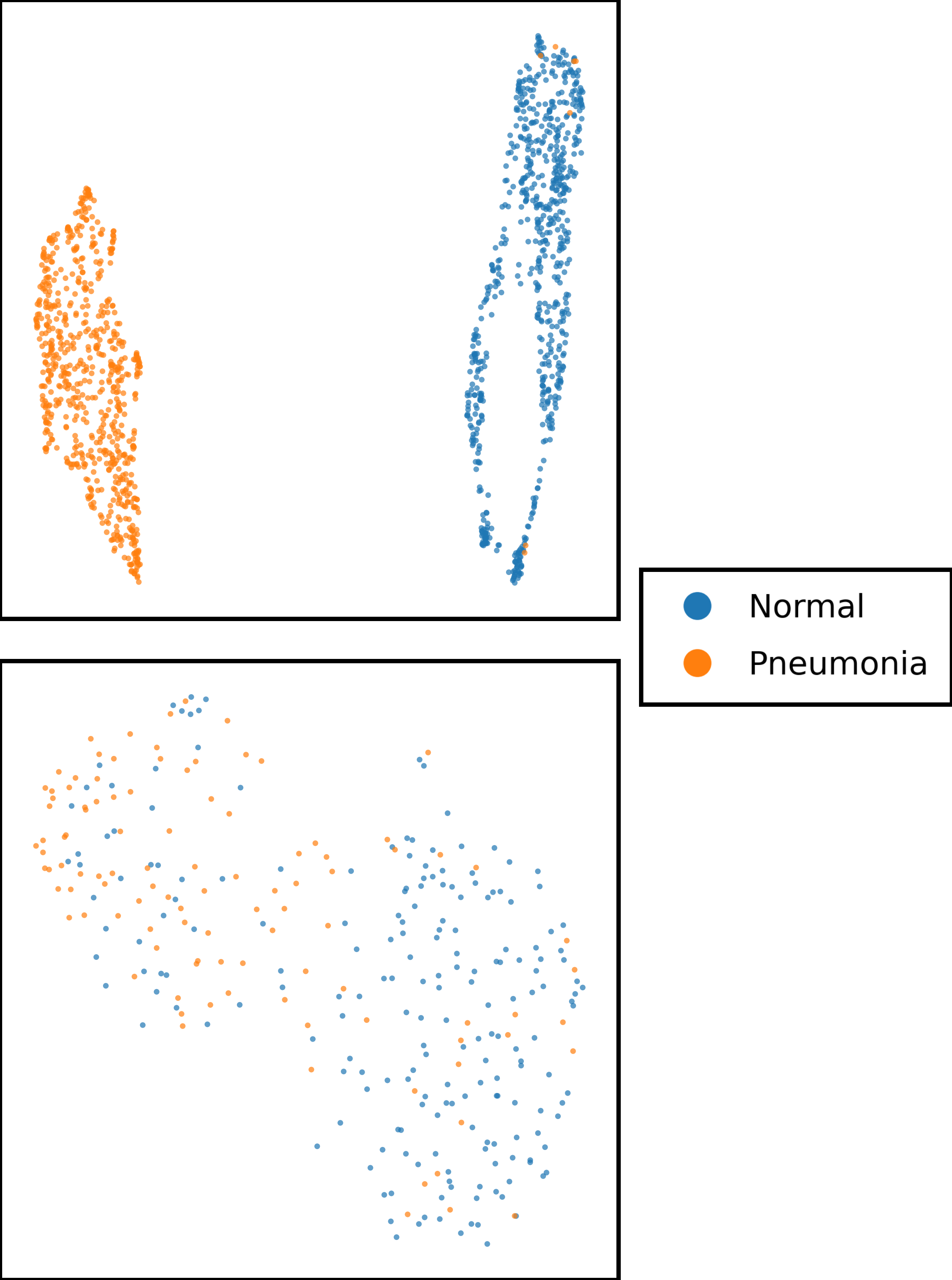}
        \caption{CT$\rightarrow$CXR}
        \label{fig:ctxr}
    \end{subfigure}
\caption{UMAP of backbone features at inference, teacher (top) and student
(bottom), colored by class, for (a) OCT$\rightarrow$fundus and (b)
CT$\rightarrow$CXR.}
    \label{fig:both}
\end{figure}

Table~\ref{tab:ablation} removes one codebook loss at a time. On
OCT$\rightarrow$fundus, all three terms contribute: removing any one lowers
F1 and $\kappa$, with $\mathcal{L}_{\mathrm{dist}}$ and
$\mathcal{L}_{\mathrm{ent}}$ having the largest effect and
$\mathcal{L}_{\mathrm{cls\text{-}cb}}$ adding class-conditional refinement.
On CT$\rightarrow$CXR, $\mathcal{L}_{\mathrm{dist}}$ remains most important
and $\mathcal{L}_{\mathrm{cls\text{-}cb}}$ continues to help, but
$\mathcal{L}_{\mathrm{ent}}$ is not beneficial: removing it leaves
performance unchanged within one standard deviation
(76.6\stdv{1.0} vs.\ 76.3\stdv{1.3} F1). With only two classes the
code marginal is far less prone to collapse, so the diversity pressure that
helps the six-class setting has little left to do; retaining
$\mathcal{L}_{\mathrm{ent}}$ costs nothing here while helping there, so we
keep one objective across both.

\subsection{Cross-Modal Alignment in the Shared Codebook}
\label{sec:bridge}

Table~\ref{tab:codebook} reports per class, the cosine similarity between
teacher and student class-mean code distributions on held-out data. Alignment
is consistently positive across all classes in both settings (means
$0.60$--$0.80$ on OCT$\rightarrow$fundus, $0.49$--$0.55$ on CT$\rightarrow$CXR),
indicating that teacher and student code usage is related rather than
independent, though we do not treat this as a precise or complete
characterization of the shared representation. Alignment is also noticeably
less stable for the smallest classes (e.g. RVO, dAMD), consistent with
$\mathcal{L}_{\mathrm{cls\text{-}cb}}$ receiving a gradient for a class only
when it co-occurs in both mini-batches, so rare classes are aligned less
consistently across training runs.

\subsection{Qualitative Visualization}
\label{sec:viz}

Figure~\ref{fig:both} visualizes within-modality class separability at
inference, computed from a single representative checkpoint. The teacher (top) forms tighter, better-separated clusters than
the student (bottom) in both settings, consistent with its role as the
upper-bound modality, though the student still shows distinguishable
class structure post-distillation. This within-modality view complements
the cross-modal codebook alignment in Sec.~\ref{sec:bridge}.

\section{Discussion}
\label{sec:discussion}

\method{} estimates class-conditional code distributions from classes that
co-occur in both teacher and student mini-batches. Consequently, rare classes
receive this signal less frequently and show less stable alignment across seeds,
as observed for RVO ($0.652\!\pm\!0.293$). This suggests
that the main limitation lies in the stability of class-wise estimates under
imbalanced data rather than in the shared-codebook mechanism itself.
Class-balanced or co-occurrence-aware sampling could make rare-class alignment
more reliable, which we leave to future work.

Another important property of \method{} is that its distillation losses operate
on aggregate distributions rather than individual samples. This is intentional:
because the teacher and student datasets are unpaired, no instance-level
correspondence should be assumed. Consequently, \method{} encourages the two
modalities to agree at the batch and class levels, through $\bar p_T$, $\bar p_S$,
$\bar p_T^{\,c}$, and $\bar p_S^{\,c}$, but it does not guarantee that any
particular teacher image and student image will be mapped to similar codes. The
method is therefore best understood as disease-level semantic transfer, not
instance-level cross-modal matching.

Finally, \method{} recovers only part of the gap between the student-only model
and the teacher upper bound: $24.3\%$ on OCT$\rightarrow$fundus and $10.0\%$ on
CT$\rightarrow$CXR. This is expected, since some teacher-side information may be
modality-specific and absent from the student image itself. For example, OCT and
CT provide structural information that fundus photography and CXR may not fully
encode. The goal of \method{} is therefore not to reproduce the teacher
performance, but to transfer the portion of disease-level structure that can be
expressed through the student modality while preserving student-only inference.

\section{Conclusion}

We presented \method{}, a shared-codebook framework for unpaired cross-modal
medical image classification. Instead of forcing teacher and student features
into direct correspondence, \method{} represents both modalities as distributions
over a common learned vocabulary and transfers knowledge by aligning these
distributions globally and class-conditionally. This design avoids
instance-level pairing, reduces reliance on direct feature-space comparability,
and represents each disease class as a multi-code semantic composition rather
than a single continuous prototype. Across OCT$\rightarrow$fundus and CT$\rightarrow$CXR classification, the same distillation mechanism consistently improves the student-only model and
outperforms the evaluated baselines. Ablation and codebook-alignment analyses
further indicate that these gains come from distribution alignment in the shared
code space. Since all teacher-side and codebook modules are discarded at
inference, \method{} adds no deployment cost. Future work will extend this
shared-codebook view to partially overlapping label spaces, unlabeled student
modalities, and rare-class-aware sampling.

\section*{Acknowledgements}
This work was supported in part by the Korea government (MSIT), IITP, under IITP-2026-RS-2020-II201821 (60\%) and RS-2019-II190421 (10\%); and by the Ministry of SMEs and Startups (MSS) under RS-2024-00514724 (30\%). Professors D.T. Le and H. Choo are also with SKAI X Inc.

\section*{Disclosure of Interest}
The authors have no competing interests to declare that are relevant to the content of this article.


%
%
\bibliographystyle{splncs04}
\bibliography{main}
\end{document}